\pdfoutput=1

\documentclass[11pt]{article}

\usepackage{acl}

\usepackage{times}
\usepackage{latexsym}
\usepackage{multicol}
\usepackage{multirow}
\usepackage{comment}
\usepackage{subfigure}
\usepackage{tabularx}
\usepackage[utf8]{inputenc} 
\usepackage[T1]{fontenc}    
\usepackage{hyperref}       
\usepackage{url}            
\usepackage{booktabs}       
\usepackage{amsfonts}       
\usepackage{nicefrac}       
\usepackage{microtype}      
\usepackage{xcolor}         
\usepackage{graphicx}
\usepackage{amsmath,amssymb}
\usepackage[capitalise]{cleveref}
\usepackage{xcolor}
\newcommand{\revisionred}[1]{\textcolor{black}{#1}}

\usepackage[T1]{fontenc}

\usepackage[utf8]{inputenc}

\usepackage{microtype}

\usepackage{inconsolata}

\title{MUSCLE: A Model Update Strategy for Compatible LLM Evolution}

\author{
\begin{tabular}{ c c c }
       Jessica Echterhoff\thanks{\noindent ~~Research conducted during an internship at Apple. Corresponding authors: jechterh@ucsd.edu,\\ mpouransari@apple.com}   &    Fartash Faghri  &    Raviteja Vemulapalli \\
    UC San Diego & Apple & Apple \\
    \end{tabular} \\
    \\
\begin{tabular}{ c c c c}
       \textbf{Ting-Yao Hu} &  \textbf{Chun-Liang Li} & \textbf{Oncel Tuzel} & \textbf{Hadi Pouransari}$^*$ \\  
    \textbf{Apple} & \textbf{Apple} & \textbf{Apple} & \textbf{Apple}\\
  \end{tabular}}

\begin{document}
\maketitle
\begin{abstract}
\revisionred{
Large Language Models (LLMs) are regularly updated to enhance performance, typically through changes in data or architecture. Within the update process, developers often prioritize improving overall performance metrics, paying less attention to maintaining compatibility with earlier model versions. Instance-level degradation (\emph{instance regression}) of performance from one model version to the next can interfere with a user's mental model \cite{bansal2019beyond} of the capabilities of a particular language model. Users having to adapt their mental model with every update can lead to dissatisfaction, especially when the new model has degraded compared to a prior version for a known use case (\emph{model update regression}).
We find that when pretrained LLM base models are updated, fine-tuned user-facing downstream task adapters experience negative flips -- previously correct instances are now predicted incorrectly. We observe model update regression between different model versions on a diverse set of tasks and models, even when the downstream task training procedures remain identical. 
We argue for the importance of maintaining \emph{model update compatibility} during updates, and present evaluation metrics designed specifically for generative tasks, while also being applicable to discriminative tasks. We propose a training strategy to minimize the extent of instance regression in model updates, involving training of a compatibility adapter that can enhance task fine-tuned language models. We show negative flips reduce by up to 40\% e.g. when updating Llama 1 to Llama 2 with our proposed method.}
\end{abstract}

\section{Introduction}
Large Language Models (LLMs) are often pre-trained on large-scale corpora to obtain a base model with general world knowledge. These base models are typically evaluated using a suite of benchmarks that mostly focus on zero/few-shot performance and in-context learning capabilities. Training these models is expensive, and only a few organizations have access to the resources needed.
Hence, to enable various user-facing applications such as summarization, chatbots, code assistants, and question-answering, practitioners often adapt pre-trained base models by training task-specific parameter-efficient adapters using downstream task datasets.

\begin{figure}
    \centering
    \includegraphics[width=0.5\textwidth]{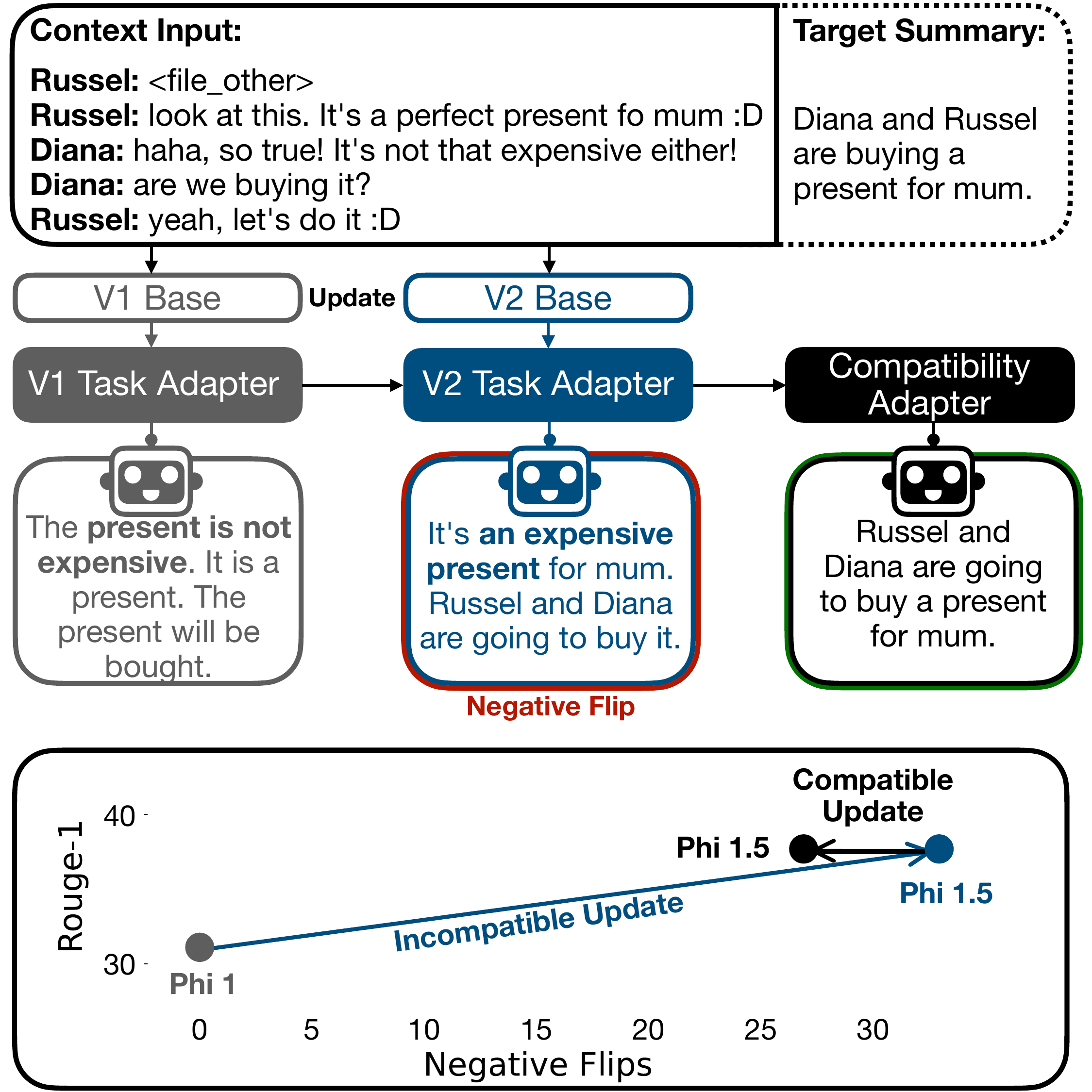}
    \caption{A real example of a model update that introduces \revisionred{instance regression (negative flip, where a previously correct prediction becomes incorrect) (top)}. With our model update strategy using a compatibility adapter approach, we enhance \revisionred{model update} compatibility to the previous model while maintaining the overall performance gain (e.g. measured by the ROUGE-1 score for the summarization task) of the model update \revisionred{(bottom)}.}
    \label{intro}
\end{figure}

Several scenarios drive updates to the base model, e.g. improved training strategy, advances in LLM architectures or increasing model context length~\cite{touvron2023llama}, the availability of additional or higher quality datasets~\cite{gunasekar2023textbooks}, the expansion of model vocabulary (e.g., to support multilingual or multimodal models), or simply training for a longer period ~\cite{biderman2023pythia}. 

\begin{figure}
    \centering
    \includegraphics[width=0.4\textwidth]{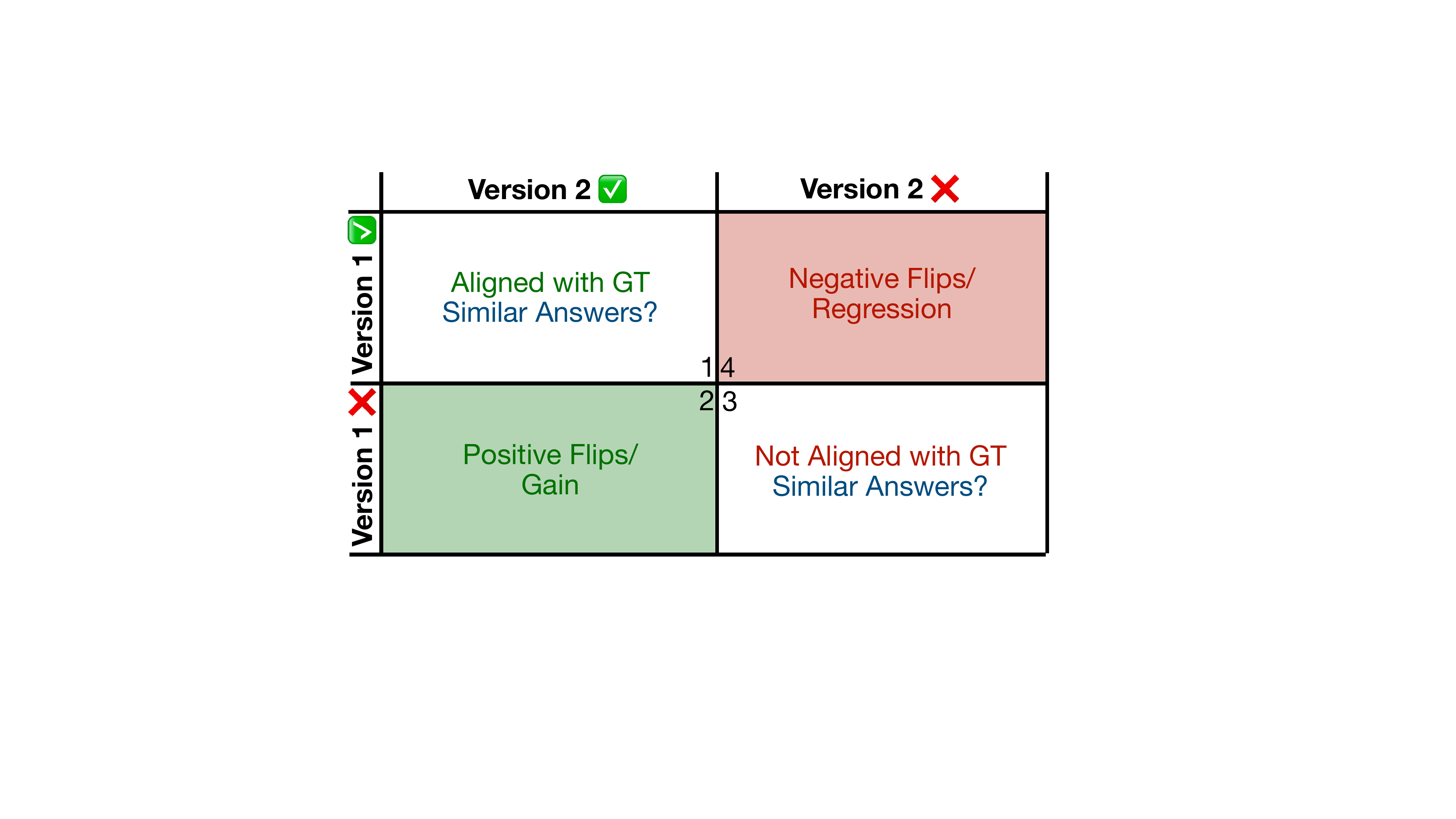}
    \caption{Four possibilities arise for each sample when a model is updated. Quadrants 2 and 4 show positive and negative flips, respectively. 
    Quadrant 3 corresponds to instances where both models are incorrect. Encouraging similarity between the old and new models in this case (i.e., making the same mistakes) results in a more seamless model update from the user's perspective.}
    \label{quadrants}
\end{figure}

When a base model is updated, all associated task adapters need to be retrained/updated to have meaningful downstream models. Hence, in the rest of the paper, we use the term \textit{model update} to refer to an update to a downstream task model, which includes updating the base model and retraining the task adapter.
\begin{table*}[]
    \centering
    \scalebox{0.95}{
     \small \begin{tabularx}{1.05\textwidth}{lX}
         \textbf{Instance Regression}  & An instance $i$ is predicted correctly by an old model $\mathcal{M}_{v1}$, but incorrectly by a new model $\mathcal{M}_{v2}$.\\
         \midrule
         \textbf{Model Update Regression} &  Aggregation of all instances correctly predicted by an old  model $\mathcal{M}_{v1}$, but incorrectly by an updated model version $\mathcal{M}_{v2}$. \\
         \midrule
         \textbf{Model Update Compatibility} & Measures the degree to which predictions made by an updated model $\mathcal{M}_{v2}$ remain consistent with the predictions of the previous model version $\mathcal{M}_{v1}$, specifically on instances where $\mathcal{M}_{v1}$ was correct. Quantifies the alignment or compatibility between model versions across updates.\\
    \end{tabularx}}
    \caption{\revisionred{Definitions of Instance Regression, Model Update Regression, and Model Update Compatibility.}}
    \label{table:definitions}
\end{table*}
When a model is updated, we evaluate \revisionred{\emph{model update compatibility} with different metrics} by four quadrants shown in \cref{quadrants}. The new model could produce a worse prediction (negative flip (quadrant 4)) for many samples~\citep{yan2021positive} even when it has a better overall performance when compared to the previous model. A real example of instance regression measured as a negative flip is shown in \cref{intro} for a dialogue summarization task. \revisionred{This kind of regression can confuse the user and impair their satisfaction~\citep{sakai2022generalized}. We denote the aggregated overall regression for all individual instances as \textit{model update regression} (Table \ref{table:definitions}). Regression testing has become increasingly important for the evolving use of LLMs accessed via APIs~\citep{ma2023my}.} 
\revisionred{When updating models, practitioners typically} focus on increasing positive flips (quadrant 2) while avoiding negative flips (quadrant 4) \cite{cai2022measuring, sakai2022generalized, yan2021positive, li2023lightweight, schumann2023backward}. However, they neglect prediction inconsistencies in the scenarios where both model versions are 
incorrect (quadrant 3) \revisionred{or already correct, but slightly different (quadrant 1).}
For example, \citet{trauble2021backward} assumes the cost of flips from one incorrect class to another incorrect class to be zero. We argue that there is value in evaluating consistency when both models are wrong. A user may have developed coping strategies on how to interact with a model when it is incorrect; therefore, inconsistencies in mistakes can lead to user dissatisfaction. 

Previous works have mainly addressed the \revisionred{model update} regression challenge for classification tasks~\cite{cai2022measuring, sakai2022generalized, yan2021positive, li2023lightweight, schumann2023backward}.
In this work, we systematically study the problem of model update compatibility 
using discriminative and generative downstream tasks and different base models.

Following is a summary of our contributions: 
\begin{itemize}
    \item We formulate the compatibility problem when updating LLMs. 
    \revisionred{We focus on common setups where base LLMs undergo updates and evaluate model compatibility and performance via parameter-efficient, task-specific fine-tuning, as these models are deployed for user interaction.}
    \item We extend the notion of model compatibility from discriminative to generative tasks, and propose compatibility metrics that consider similarities in model behavior after an update, going beyond the negative flip rate metric.
    \item We investigate model compatibility for different update scenarios using open-weight models and find significant \revisionred{model update} regression across various tasks.
    \item We propose learning a compatibility adapter to align model versions and minimize \revisionred{model update regression}. We demonstrate up to 40\% reduction in negative flip rate (e.g. for Llama 1 to Llama 2 update in language understanding) and reduced model inconsistency for downstream tasks such as summarization, math reasoning, and commonsense question-answering.
 \end{itemize}

\section{Related Work}
\subsection{Measuring \revisionred{Model Update} Regression}
\paragraph{Classification}
\citet{yan2021positive} introduce  negative flip rate (\text{NFR}) to evaluate model compatibility for classification tasks. \text{NFR} calculates the fraction of instances that were previously correct but are now incorrect with the new model. A similar statistic~\cite{sakai2022generalized}, backward trust compatibility (BTC)~\citep{srivastava2020empirical}, measures the ratio of instances that the new model predicts correctly among all instances the old model predicts correctly.  \citet{matsuno2023robust} propose backward compatibility with a conditional distribution, which computes the ratio at which the accuracy of the conditional distribution of the new model is equal to or higher than that of the old model. \citet{cai2022measuring} introduce the negative flip impact for graph NLP tasks, taking into account the negative flips and the overall error rate of the model. These aforementioned metrics are limited to the evaluation of discriminative classification tasks.

\subsection{Reducing \revisionred{Model} Update Regression}
\paragraph{Model Ensembles} Prior work found that model ensembles reduce model update regression. This can be attributed to the reduction of variance by ensembling, as every single model may capture the training data from a distinct aspect~\citep{yan2021positive, xie2021regression}. 
Extensions on this line of work include choosing the most centric model from an ensemble~\cite{xie2021regression}, aligning two models' uncertainties \cite{li2023lightweight}, or using gating mechanisms~\cite{lai2023improving}. 
Previous work has also used model parts from an old model to infuse into the new one~\citep{ran2023xadapter}, with the limitation of both models being required at inference time. For limited use cases, \citet{qin2023recyclable} have shown to re-use previously learned adapters when purely a data update was performed. All these methods either introduce a larger memory footprint by re-using old model parts or are limited to (same-domain) data-updates and same models.
\paragraph{Knowledge Distillation}
Originally proposed for model compression \cite{buciluǎ2006model}, a (smaller) student model is trained to mimic a (larger) teacher model. By treating the old model as the teacher and the new model as the student, knowledge distillation has been shown to reduce model update regression in vision and language discriminative tasks \cite{yan2021positive,xie2021regression,schumann2023backward,jaeckle2023fastfill,zhang2021hot,bct,fct}. \citet{bct} propose a distillation-based influence loss to align new model representations with those of the old model. Similarly, \citet{fct,jaeckle2023fastfill} apply distillation after a learned transformation module. \citet{schumann2023backward} propose weight interpolation between the old and the new model, \citet{zhao2022elodi} suggest matching old and new model distributions, \citet{yan2021positive} distill from an ensemble, and \citet{caciolai2023regression} recommend using focal distillation. However, none of these approaches evaluate their approaches on generative tasks or they require both the old and new models to be in memory at inference time.

\section{Problem Formulation} 
\revisionred{While existing methods provide valuable approaches to mitigating model update regression, they predominantly focus on discriminative classification tasks and often require additional memory at inference time. We propose a flexible approach to updating models without sacrificing performance or compatibility across downstream tasks.}
\paragraph{Setup}
We follow the common setup of finetuning a pre-trained base LLM to multiple downstream tasks with task-specific LoRA adapters~\citep{hu2021lora}. Let $\mathcal{M}^{\text{base}}_i$ denote the $i^{th}$ version of a base LLM with parameters $\theta_i$. We adapt $\mathcal{M}^{\text{base}}_i$ to a downstream task $\mathcal{T}$ using an adapter $\mathcal{A}_{i}^\mathcal{T}$ to obtain a downstream model $\mathcal{M}_i^{\mathcal{T}}$ with weights $\theta^\mathcal{T}_i=\theta_i + \Delta_i^\mathcal{T}$, where $\Delta_i^\mathcal{T}$ denotes the weights of the task-specific adapter $\mathcal{A}_i^\mathcal{T}$ learned using the training data corresponding to task $\mathcal{T}$. When the base model is updated from $\mathcal{M}^{\text{base}}_{v1}$ to $\mathcal{M}^{\text{base}}_{v2}$, the task-specific adapters are re-trained for each downstream task. Hereafter, for simplicity of notation, we use $\mathcal{M}_{v1}$ and $\mathcal{M}_{v2}$ to refer to the task-adapted models $\mathcal{M}_{v1}^\mathcal{T}$ and $\mathcal{M}_{v2}^\mathcal{T}$, respectively, and explicitly mention the task $\mathcal{T}$ when needed.
\subsection{Backward Compatibility Metrics}
A backward compatibility metric outputs a compatibility score based on two models, $\mathcal{M}_{v1}$ to $\mathcal{M}_{v2}$. 
\citet{yan2021positive} propose negative flip rate (\text{NFR}) to measure regression in classification models over a dataset $\{x_i, y_i\}_{i=1}^N$, where $y_i$ is the ground truth class for input $x_i$ for a particular task $\mathcal{T}$:
\begin{equation*}
\text{NF}(x_i) \triangleq[\mathcal{M}_{v1}(x_i) = y_i] \land  [\mathcal{M}_{v2}(x_i) \neq y_i]
\end{equation*}
\begin{equation*}
    \text{NFR} \triangleq \frac{1}{N} \sum_i^N \mathbf{1}[\text{NF}(x_i)]
\end{equation*}
Here $\mathbf{1}$ denotes the indicator function. This notion of regression is partly applicable to autoregressively trained tasks. LLM benchmarks~\cite{eval-harness} calculate the likelihood of every possible choice in a multiple-choice question and choose the response with the highest likelihood to calculate \revisionred{a \emph{likelihood based accuracy}}. This evaluation can indicate \revisionred{model update} regression similar to classification for tasks when multiple choices are available~\cite{zellers2019HellaSwag, wang2019superglue, welbl2017crowdsourcing}. 

\paragraph{Unobserved Inconsistencies} \revisionred{Other inconsistencies arise when the old model predicts class A, the new model class B, and the ground truth is class C. Similarly, if there are multiple ground truth options, a flip could occur within the ground truth options, as in quadrant 1 in \cref{quadrants}. These inconsistencies are not captured by positive or negative flips. We argue that if we cannot produce a better prediction, we should at least stay consistent with the user's expectations and propose extended negative flips metrics for this use case.}
\begin{figure*}[t]
    \centering
    \includegraphics[width=0.9\textwidth]{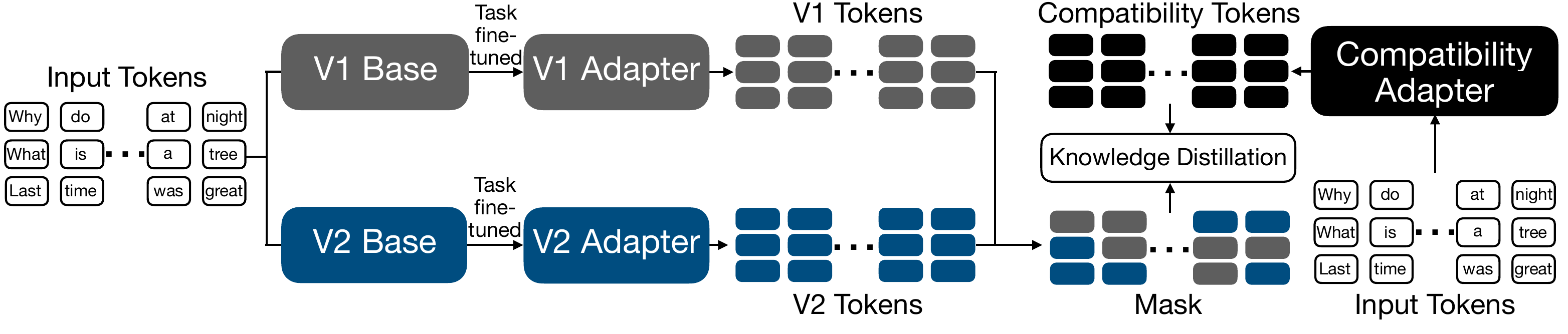}
    \caption{When updating a model, regression on individual tokens and instances can arise. We use a masked approach to select tokens to be aligned with knowledge distillation either with the old version to remain consistent or with the new task model to increase performance.}
    \label{fig:flow}
\end{figure*}

\paragraph{Continuous Metrics}
In generative tasks for language models, we do not necessarily have multiple choices for which we can predict if an instance regressed or not. Hence a metric incorporating multiple choices and then calculating negative flips is not applicable for continuous evaluation metrics. \revisionred{Typical continuous metrics in generative language tasks are ROUGE~\citep{lin2004ROUGE} or BERT~\citep{zhang2019bertscore} scores. As regular negative flips are incapable of capturing these nuances,} we require a new metric to evaluate compatibility for generative tasks like summarization.
\subsection{Extended Evaluation Metrics}\label{sec:metrics}
\revisionred{We propose a suite of metrics that evaluate model update compatibility on a fine-grained basis specifically for generative tasks (e.g summarization).}
\paragraph{\revisionred{Accounting for Flips when Both Models are Incorrect}}
To capture inconsistencies for instances where both old and new models are incorrect, we adapt the negative flip rate as follows when multiple choice options are available: 
\begin{align*}
   \text{NF}_{mc}(x_i) \triangleq &[\mathcal{M}_{v2}(x_i) \neq y_i ] \land \\ &[\mathcal{M}_{v1} (x_i) \ne \mathcal{M}_{v2}(x_i)]
\end{align*}
\begin{align*}
    \text{NFR}_{mc} \triangleq\frac{1}{N} \sum_i^N \mathbf{1}[
     \text{NF}_{mc}(x_i)]
\end{align*}
This includes the possibility that neither of the models gives the correct answer, but a change in behavior occurs that can confuse a user. 
Similarly, for multi-label tasks, this notion can account for ground-truths that may have flipped during a model update when multiple ground truth options exist.

\paragraph{\revisionred{Smooth Compatibility Metrics}}
To add continuous metrics for generative tasks, we evaluate the \emph{expected} \revisionred{model update regression}.
Our general framework aims to be independent of the actual similarity metric, such that it can be chosen in accordance with any respective task of interest. \revisionred{For example, translation tasks might require BLEU~\citep{papineni2002bleu}, but summarization ROUGE~\citep{lin2004ROUGE} evaluation. Additionally, we want to measure a notion of \emph{performance gain}, when both models are correct, but one might still be better than the other.}

Given a similarity metric $S$, and model outputs $\mathcal{M}_{v1}(x_i)$ and $\mathcal{M}_{v2}(x_i)$ for an input $x_i$, the difference for a model update for a particular test instance $i$ is
\begin{equation*}
    D(x_i) \triangleq S(\mathcal{M}_{v2}(x_i),y_i)-S(\mathcal{M}_{v1}(x_i),y_i)
\end{equation*}\label{eq:distance}
acting as an indication of the distance between the two model outputs with respect to the grand truth.  This per instance indication of distance enables us to classify which model update compatibility quadrant an instance falls into. 
In practice, similarity metrics could be BERT Score \cite{zhang2019bertscore}, ROUGE Score \cite{lin2004ROUGE}, BLEU Score \cite{papineni2002bleu}, or model-as-a-judge metrics \cite{huang2024empirical} depending on the use case and task. 

To get a \revisionred{metric for generative tasks} that is similar to the positive and negative flip rate in \revisionred{classification} tasks, we observe the distribution of instances with positive gain or negative regression:
\begin{equation*}
    \widetilde{\text{PFR}} \triangleq \frac{1}{N}\sum_{i}^N \mathbf{1}[D(x_i) > 0]
\end{equation*}
\begin{equation*}
    \widetilde{\text{NFR}} \triangleq \frac{1}{N}\sum_{i}^N \mathbf{1}[D(x_i) < 0]
\end{equation*}

To observe the magnitude of \revisionred{model update} regression and gain, we can observe the expectation:

\begin{equation*}
    m_g \triangleq \frac{1}{N \cdot \widetilde{\text{PFR}}}\sum_{i}^N D(x_i)~\mathbf{1}[D(x_i) > 0]
\end{equation*}
\begin{equation*}
    m_r \triangleq \frac{1}{N \cdot \widetilde{\text{NFR}}}\sum_{i}^N |D(x_i)|~\mathbf{1}[D(x_i) < 0]
\end{equation*}
These give an indication of the average magnitude of change in a similarity metric \revisionred{when gain or regression occurs. }

\begin{table*}
\centering
\scalebox{0.95}{
\small
\centering
\begin{tabularx}{\textwidth}{l|l|X}
     $\mathcal{M}^{\text{base}}_{v1}$ & $\mathcal{M}^{\text{base}}_{v2}$ & Update  \\
     \midrule
     Phi 1 & Phi 1.5 &  Synthetic data, data selection \cite{gunasekar2023textbooks, li2023textbooks}\\
     Phi 1.5 & Phi 2 &  Synthetic data, data selection, $\uparrow$ parameters \cite{javaheripi2023phi}\\
     Llama 1& Llama 2 & More data, $\uparrow$ context size \cite{touvron2023llama, touvron2023llama2} \\
    Vicuna 1.3 & Vicuna 1.5 & Llama 1 $\rightarrow$ Llama 2, Instruction fine-tuning \cite{zheng2024judging}\\
\end{tabularx}}
\caption{We select a suite of models with varying update scenarios and parameter sizes and evaluate them on different LLM benchmark tasks. We select Llama and Vicuna models with 7B parameters.}
\label{tab:model_updates}
\end{table*}

\begin{table}
\scalebox{0.91}{
\small
\begin{tabularx}{0.53\textwidth}{l|c|c}
     Dataset & Task   &  \revisionred{Metric}\\
     \midrule
     HellaSwag & 
     \begin{tabular}{c}
     Language Understanding\\
     (Multiple-Choice) 
     \end{tabular} &\begin{tabular}{c} \revisionred{Log-Likelihood} \\ \revisionred{Accuracy} \end{tabular}
     \\ 
     PIQA & 
     \begin{tabular}{c}
     Commonsense Reasoning\\
     (Binary-Choice)
     \end{tabular} & \begin{tabular}{c} \revisionred{Log-Likelihood} \\ \revisionred{Accuracy} \end{tabular}
     \\  
     GSM8k & 
     \begin{tabular}{c}
     Grade School Math\\
     (Exact Match)
     \end{tabular} & \begin{tabular}{c} \revisionred{Exact Match} \\ \revisionred{Accuracy} \end{tabular}
     \\
     SAMsum & Dialogue Summarization & \revisionred{ROUGE-1}
\end{tabularx}
}\caption{We tackle compatible model updates for different downstream tasks and datasets, including multiple-choice and generative tasks~\citep{zellers2019HellaSwag, DBLP:journals/corr/abs-1911-11641, DBLP:journals/corr/abs-2110-14168, gliwa2019samsum}.
}\label{tab:tasks}
\end{table}

\section{Knowledge Transfer}
\revisionred{Now that we have metrics to indicate model update regression, we propose a knowledge distillation approach to minimize this regression for the task-specific models $\mathcal{M}_{v1}$ and $\mathcal{M}_{v2}$}. Typically, knowledge distillation minimizes the KL divergence between the soft targets $\sigma(z_t)$ and $\sigma(z_s)$, where $z_s$ and $z_t$ are logits predicted by student and teacher models, respectively. 
\begin{equation*}
    \mathcal{L}_{KL} = \frac{1}{n}\sum_{i=1}^n KL(\sigma(z_{t,i}/T)\| \sigma(z_{s,i}/T))
\end{equation*}
$,i$ denotes the $i$'th token, $n$ is the total number of tokens available for training, $T$ is temperature parameter, and $\sigma$ denotes softmax.
Most knowledge distillation works consider the distillation from a trained teacher to an untrained student \cite{tian2022contrastive, rajasegaran2020self}. Recent work~\citep{roth2024fantastic} tackles the goal of knowledge transfer between pre-trained student and teacher models while retaining student knowledge gained a priori, and shows that standard knowledge distillation between pre-trained models struggles to transfer knowledge without performance drops.
Complementary to this work focusing on performance and maintaining prior knowledge, we tackle compatibility with prior models through knowledge transfer. 

\subsection{Model Update Strategy for Compatible LLM Evolution (MUSCLE)} \label{sec:muscle}
When the base model is updated, we train a task-specific fine-tuned model, $\mathcal{M}^{C}_{v2}$, that has the accuracy benefits of $\mathcal{M}_{v2}$, but with most compatibility with $\mathcal{M}_{v1}$. We obtain $\mathcal{M}^{C}_{v2}$ by training a \emph{compatibility adapter} applied to the base model $\mathcal{M}^{\text{base}}_{v2}$. 
We use knowledge from task-specific fine-tuned models $\mathcal{M}_{v1}$ and $\mathcal{M}_{v2}$ when training $\mathcal{M}^{C}_{v2}$. $\mathcal{M}_{v2}$ typically has increased prediction capabilities over $\mathcal{M}_{v1}$ (due to improvements in the base model), but $\mathcal{M}_{v1}$ has information on already correctly predicted tokens or instances that we want to minimize \revisionred{degradation} towards.

We initialize the compatibility adapter with the task-specific adapter of $\mathcal{M}_{v2}$, and further fine-tune it (using the task training dataset) by aligning the next token prediction to either $\mathcal{M}_{v1}$ or $\mathcal{M}_{v2}$.
We define masking (for individual tokens of a training sequence) following a simple heuristic depending on whether $\mathcal{M}^{C}_{v2}$ (the adapter being trained) predicts the correct tokens or not. If it does, we align to  $\mathcal{M}_{v2}$ logits, otherwise, we align to $\mathcal{M}_{v1}$. The fine-tuning process of $\mathcal{M}^{C}_{v2}$ is  depicted in \cref{fig:flow}. The fine-tuning loss to train the compatibility adapter, $\mathcal{L}_{comp}^m$, is defined below:

\begin{equation*}
 m_i = \mathbf{1}[\text{argmax}~\sigma(z_{\mathcal{M}^C_{v2},i}) \neq y_i]
\label{masking}
\end{equation*}
\begin{align}
\begin{split}
    a_{\mathcal{M}_{v1}} =& KL(\sigma(z_{\mathcal{M}_{v1},i}/T)\| \sigma(z_{\mathcal{M}^C_{v2},i}/T)) \\
    a_{\mathcal{M}_{v2}} = &KL(\sigma(z_{\mathcal{M}_{v2},i}/T) \| \sigma(z_{\mathcal{M}^C_{v2},i}/T)) \\
    \mathcal{L}_{comp}^m =& \frac{1}{n} \sum_{i=1}^n m_i \cdot a_{\mathcal{M}_{v1}}
    + (1-m_i) \cdot a_{\mathcal{M}_{v2}}
\end{split}
\label{eqn:loss}
\end{align}

When evaluating, we denote $\text{NFR}$ as negative flip rate between $\mathcal{M}_{v1}$ $\mathcal{M}_{v2}$, $\text{NFR}_c$ as the observed negative flip rate between $\mathcal{M}_{v1}$ and our compatibility model $\mathcal{M}^{C}_{v2}$, and $\Delta \text{NFR}_c = \text{NFR}_c - \text{NFR}$.

\begin{table*}[t]
\small
\centering
\renewcommand{\arraystretch}{1.1}
    \begin{tabular}{l|ll|rr|rr|rrr}
\toprule
& $\mathcal{M}^{\text{base}}_{v1}$ & $\mathcal{M}^{\text{base}}_{v2}$ & $\text{acc}_{\mathcal{M}_{v1}}$ & $\text{acc}_{\mathcal{M}_{v2}}$ & $\text{acc}_{c}$ & $\Delta \text{acc}_{c}\uparrow$ & $\text{NFR}$ & $\Delta \text{NFR}_c\downarrow$ & $\Delta_{\%} \text{NFR}_c\downarrow$\\
\midrule
\multirow{4}{*}{\rotatebox[origin=c]{90}{\parbox{1.4cm}{\centering HellaSwag}}} & Llama 1 & Llama 2 & 72.74 & 72.91 & 79.53 & 6.62 & 10.27 & -4.17 & \textbf{-40.60} \\
& Phi 1.5 & Phi 2 & 69.76 & 78.02 & 77.86 & -0.16 & 3.09 & 0.24 & 7.77  \\
& Phi 1 & Phi 1.5 & 32.38 & 69.76 & 69.86 & 0.10 & 2.47 & -0.13 & \textbf{-5.26}  \\
& Vicuna 1.3 & Vicuna 1.5 & 72.19 & 71.54 & 78.10 & 6.56 & 10.48 & -4.06 & \textbf{-38.74} \\

\midrule
\multirow{4}{*}{\rotatebox[origin=c]{90}{\parbox{1cm}{\centering PIQA}}} & Llama 1 & Llama 2 & 74.86 & 74.76 & 79.27 & 4.51 & 11.59 & -3.97 & \textbf{-34.25} \\
& Phi 1.5 & Phi 2 & 74.65 & 78.78 & 78.94 & 0.16 & 6.20 & -0.11 & \textbf{-1.77} \\
& Phi 1 & Phi 1.5 & 59.03 & 74.65 & 74.70 & 0.05 & 7.78 & -0.05 & \textbf{-0.64}  \\
& Vicuna 1.3 & Vicuna 1.5 & 74.70 & 75.52 & 78.89 & 3.37 & 9.90 & -2.88 & \textbf{-29.09} \\
\bottomrule
\end{tabular}\caption{Compatible task adapter trained with MUSCLE (corresponding to metrics with suffix $c$) reduces negative flip rate \revisionred{on the test sets of multiple-choice language tasks}. We see most improvements for model updates that have small performance differences.}\label{classificationtable}
\end{table*}
\begin{table*}[]
\small
\centering
\renewcommand{\arraystretch}{1.1}
\begin{tabular}{ll|rr|rr|rrr}%
\toprule
 $\mathcal{M}^{\text{base}}_{v1}$ & $\mathcal{M}^{\text{base}}_{v2}$  & $\text{EM}_{\mathcal{M}_{v1}}$ & $\text{EM}_{\mathcal{M}_{v2}}$ & $\text{EM}_c$ & $\Delta \text{EM}_c\uparrow$ & $\text{NFR}$ & $\Delta \text{NFR}_c\downarrow$& $\Delta_{\%} \text{NFR}_c\downarrow$\\
\midrule
Llama 1 & Llama 2 & 24.45 & 33.09 & 36.66 & 3.57 & 8.49 & -0.91 & \textbf{-10.72} \\
Phi 1.5 & Phi 2 & 30.02 & 48.18 & 50.68 & 2.50 & 5.88 & -1.71 & \textbf{-29.08} \\
Phi 1 & Phi 1.5 & 3.41 & 30.02 & 26.99 & -3.03 & 2.01 & -0.04 & \textbf{-1.99}  \\
Vicuna 1.3 & Vicuna 1.5 & 26.72 & 29.91 & 31.84 & 1.93 & 11.60 & -0.99 & \textbf{-8.53} \\
\bottomrule
\end{tabular}\caption{GSM8K math evaluation with exact match (EM) \revisionred{over the test dataset}. For compatibility adapter (corresponding to metrics with suffix $c$), we observe a decreased negative flip rate while mostly maintaining performance gains. }
\label{gsm8k}
\end{table*}
\section{Experimental Setup}
\subsection{Model Update Assumptions}
To analyze the impact of model updates we consider parameter-efficient fine-tuned models using Low-Rank Adapters (LoRA) \cite{hu2021lora}. Compared to previous work on continuous learning and model updates~\cite{qin2022elle, qin2023recyclable}, we do not limit model updates to be produced by only data updates, but consider different kinds of updates shown in \cref{tab:model_updates}. We include updates due to data, increased parameters, or different training strategies. We include a wide range of downstream tasks to evaluate model compatibility, including \emph{generative tasks} as summarized in \cref{tab:model_updates}.
For all tasks, we learn the LoRA adapter autoregressively (next-token prediction). 

\subsection{Task Adapter Training}
For each task and each old model $\mathcal{M}_{v1}$ and new model $\mathcal{M}_{v2}$, we train a LoRA adapter on all linear layers with $r=128$ and $\alpha=256$. We use a 0.8/0.2 training/validation split for 10 epochs. We choose the model by cross-entropy validation loss. More information on hyperparameters is shown in \cref{tab:hyperparams}.

\subsection{Compatibility Adapter Training}
We keep all hyper-parameters the same for task adapter training, and train compatibility adapter with the $\mathcal{L}_{comp}^m$ loss defined in \cref{eqn:loss}. We analyze the statistical significance of the results on a subset of compatibility adapter training for Phi 1 to Phi 1.5 updates on PIQA using 3 random seeds. We observe a standard deviation of accuracy of 0.0012.

We compare $\mathcal{L}_{comp}^m$ with different masking strategies $m$ in the ablation studies in \cref{sec:masking}, but find that the version introduced in \cref{sec:muscle} works best for most model updates. For model updates with large performance gaps between $\mathcal{M}_{v1}$ and $\mathcal{M}_{v2}$ we find that an auxiliary cross-entropy loss enhances the stability of training for PIQA and SAMSum from Phi 1 to Phi 1.5, and all Phi updates in GSM8k and Hellaswag (more in \cref{celoss}).

\subsection{Similarity Metrics}
\revisionred{ We evaluate multiple-choice tasks with classification-like approaches by choosing the maximum log-likelihood of the possible answers~\citep{eval-harness}.
For math tasks, we use exact-match accuracy of the final calculated result. For summarization, we cannot evaluate in a classification-like manner. We use ROUGE-1 ~\citep{lin2004ROUGE}, given that we do not observe relative ranking differences for different ROUGE-n.}

\section{Results}
\subsection{Negative Flips Occur in Model Updates}
\begin{figure}
    \centering
    \includegraphics[width=0.81\linewidth]{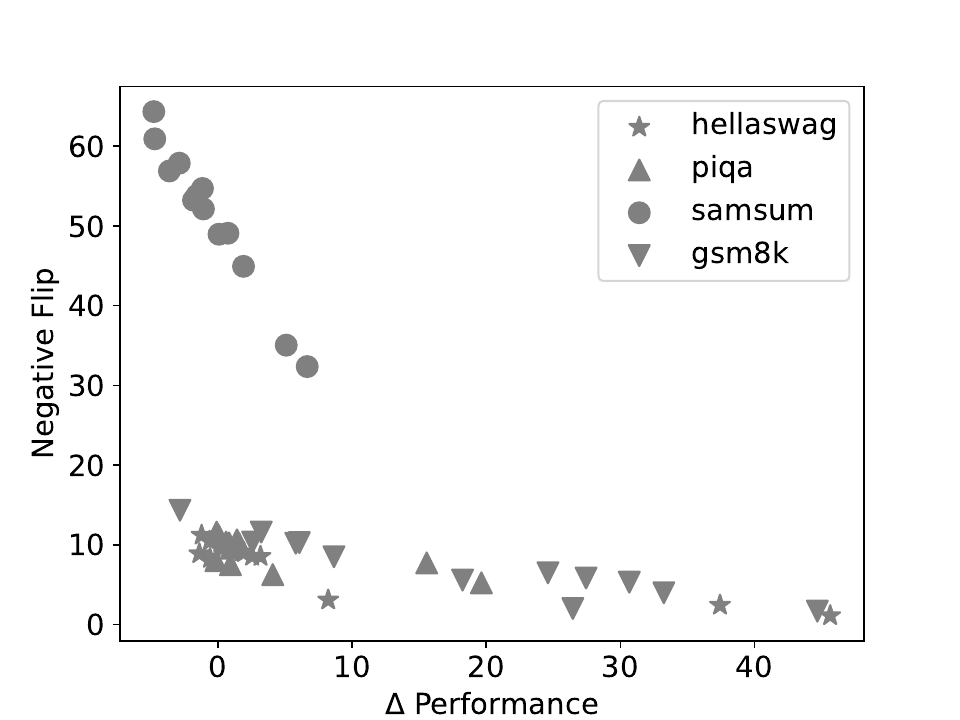}
    \caption{When updating LLM models (e.g. Llama 1 $\rightarrow$ Llama 2), we observe negative flips in different tasks. The smaller the performance gap from an old model to a new model, the more negative flips we observe.\revisionred{ We indicate the performance gap by the difference in exact match for GSM8K, Rouge-1 for SAMSum, and log-likelihood-based accuracy for PIQA and HellaSwag.} When evaluating continuous metrics with absolute ROUGE-1 value for summarization on SAMSum, we observe a large fraction of negative flips. We show the exact models analyzed in \cref{tab:mp}.}
    \label{motivation}
\end{figure}

\cref{motivation} shows that significant negative flips (up to more than 60\%) exist for a variety of base model update scenarios and downstream tasks. We observe negative flips in updates within one model family (e.g. Llama/Vicuna, and Phi models). We find more negative flips for model updates with a smaller delta in performance gain. For generative tasks like SAMsum dialogue summarization, we observe a large number of negative flips
as continuous metrics are more sensitive to small changes when updated. 
\subsection{Reduced Negative Flips in Classification}
In \cref{classificationtable}, we observe that MUSCLE decreases negative flips when compared to regular model updates without compatibility-specific training ($\mathcal{M}_{v2}$). Specifically, we see a reduction of NFR by 40\% for the update of LLama 1 to Llama 2 and 39\% for Vicuna 1.3 to Vicuna 1.5. For updates with a large performance gap (for example Phi 1 to Phi 1.5), we observe a less strong enhancement with a negative flip rate reduction by 1-5\%. In addition to the reduction of negative flips, we also observe an increased accuracy of up to 7\% for Llama and Vicuna updates. 

In exact match (EM) evaluation, we match the final result of the math question from the prediction to ground truth. Results are shown in \cref{gsm8k}. In this case, we observe that we can reduce the number of negative flips by 29\% for Phi 1.5 to Phi 2. When the version 1 model is significantly less accurate \revisionred{(e.g., 3.4\% exact-match accuracy for Phi 1), we observe a reduction in accuracy with the compatibility adapter} while only being able to decrease negative flips by 2\%. For all other updates, MUSCLE increases expected-match accuracy while reducing negative flips. 

\begin{table*}[]
    \centering
    \small
    \renewcommand{\arraystretch}{1.1}
\resizebox{\textwidth}{!}{
\begin{tabular}{ll|rr|rr|rrr|rr}
\toprule
 $\mathcal{M}^{\text{base}}_{v1}$ & $\mathcal{M}^{\text{base}}_{v2}$ & $\text{R1}_{\mathcal{M}_{v1}}$ & $\text{R1}_{\mathcal{M}_{v2}}$ & $\text{R1}_c$ & $\Delta \text{R1}_c\uparrow$ & $\widetilde{\text{NFR}}$ & $\Delta \widetilde{\text{NFR}}_c\downarrow$ & $\Delta_{\%}\widetilde{\text{NFR}}_c\downarrow$& $\Delta m_{g}\uparrow$ & $\Delta m_{r}\downarrow$ \\
 \midrule
 Llama 1 & Llama 2 & 32.06 & 32.28 & 34.79 & 2.51 & 48.96 & -7.81 & \textbf{-15.95} & 0.64 & \textbf{-1.13} \\
Phi 1.5 & Phi 2 & 37.53 & 36.15 & 40.69 & 4.54 & 54.70 & -15.02 & \textbf{-27.46} & 0.24 & \textbf{-3.39} \\
Phi 1 & Phi 1.5 & 30.92 & 37.53 & 38.76 & 1.23 & 32.60 & -5.74 & \textbf{-17.61} & -0.38 & \textbf{-0.59} \\
Vicuna 1.3 & Vicuna 1.5 & 30.32 & 30.88 & 34.08 & 3.20 & 49.69 & -10.98 & \textbf{-22.10} & -0.02 & \textbf{-1.85} \\
\bottomrule
\end{tabular}
}
    \caption{For summarization (SAMsum) generative task we reduce \revisionred{model update} regression of ROUGE-1 score (R1) by up to 27\%.}
    \label{tab:samsum}
\end{table*}

\subsection{Increased Consistent Behavior}
\begin{figure}
    \centering
    \includegraphics[width=\linewidth]{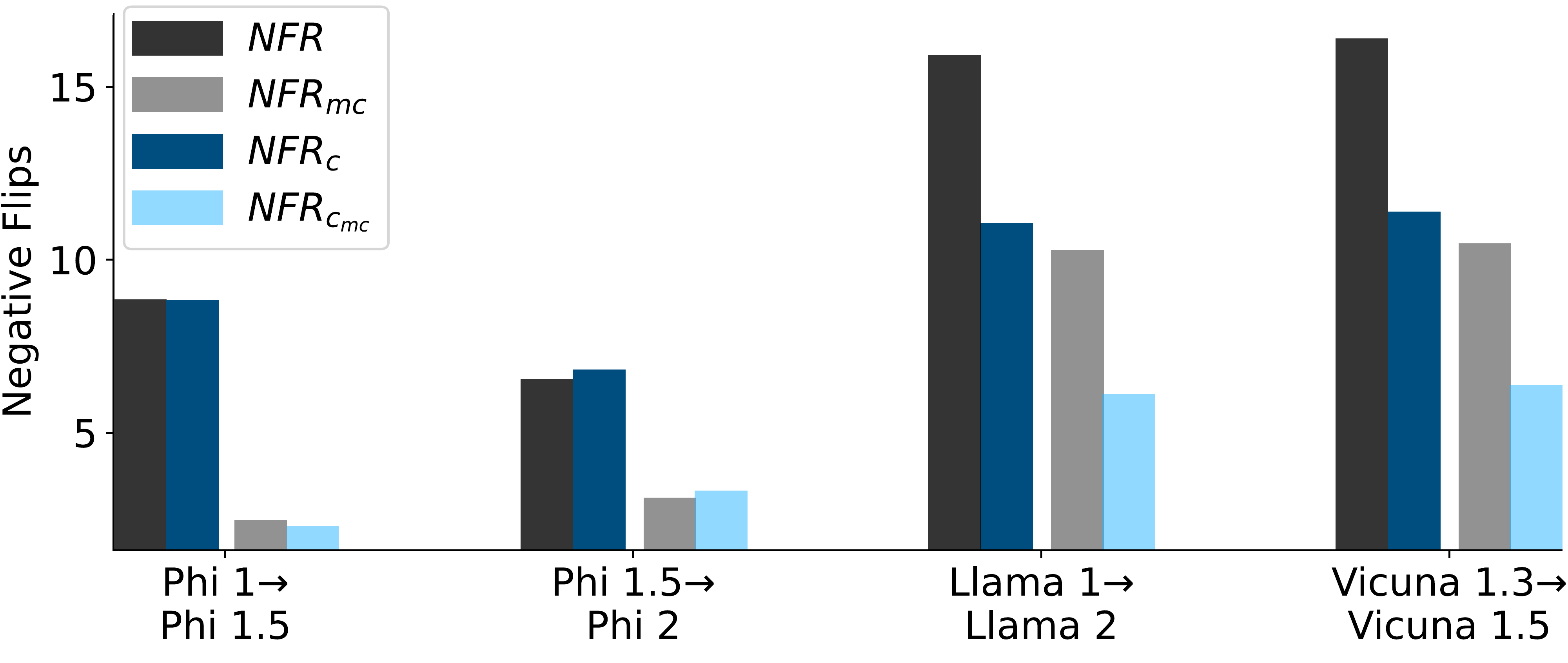}
    \caption{Comparison of NFR vs NFR$_{mc}$ metrics to evaluate inconsistency when updating LLMs for HellaSwag task. \revisionred{We see that using our compatibility adapter (denoted by $_c$), we can reduce inconsistency for Llama and Vicuna models.}}
    \label{fig:addedNFs}
\end{figure}

When we cannot achieve a positive flip (switching from an incorrect to a correct answer), we might prefer to at least maintain consistent behavior to the old model to avoid unexpected behavior for the user. We evaluate negative flips rate \revisionred{($\text{NFR}$) and inconsistency flips rate ($\text{NFR}_{mc}$)}. For model updates that have a large performance gap and a small number of negative flips to begin with (Phi models), we see a limited reduction in inconsistency flips. However, we observe that we can reduce the inconsistency flips with our method for model updates with small accuracy gaps such as the updates for Llama and Vicuna (\cref{fig:addedNFs}) on the HellaSwag dataset. 

\subsection{Reduced \revisionred{Model Update} Regression in Generative Tasks}
When evaluating generative tasks, we can reduce \revisionred{model update} regression of ROUGE-1 score performance (\cref{tab:samsum}). We can reduce negative flips by 18\% for updates with weaker version 1 models (Phi 1 $\rightarrow$ Phi 1.5) and 22\% for smaller model updates (Vicuna 1.3 $\rightarrow$ Vicuna 1.5) and 27\% for Phi 1.5 $\rightarrow$ Phi 2. We see that we decrease ROUGE-1 \revisionred{model update} regression by 1-3\%, while maintaining gain.
\subsection{The Effect of Different Masking Strategies}
\label{sec:masking}
We analyze different training and masking strategies to evaluate our design choices. On the PIQA dataset and model update Llama 1 $\rightarrow$ Llama 2, we compare MUSCLE with different masking strategies.
Intuitively, \revisionred{instance or token-wise} likelihood-based masking strategies can be useful for tasks that are evaluated with log-likelihoods (e.g. PIQA, HellaSwag) where multiple choices are available. We compare likelihood-based masking for example on individual tokens, 
\begin{equation}
    m_i = LL_L = \mathbf{1}[\sigma(z_{\mathcal{M}^C_{v2},i})<\sigma(z_{\mathcal{M}_{v1},i})]
\end{equation}\label{eqn:LLbased} to check if the likelihood of the ground-truth next token
in the current model is smaller than the old model. Only in this case, we align to $\mathcal{M}_{v1}$, and align to $\mathcal{M}_{v2}$ for every other token. 
Alternatively, we can also compare the likelihood of the entire sequence 
\begin{equation}
    m_i = LL_S = \mathbf{1}[\sum_i[\sigma(z_{\mathcal{M}^C_{v2},i})<\sigma(z_{\mathcal{M}_{v1},i})]]
\end{equation}\label{equation:LLS}
such that we mask all tokens per sequence to get instance-wise masking. For both of these strategies, we see a reduction in negative flips. We note that likelihood-based masking requires auxiliary cross-entropy loss for stability (\ref{celoss}). 
\begin{table}
    \small
    \centering
    \renewcommand{\arraystretch}{1.2}
\resizebox{\columnwidth}{!}{
    \begin{tabular}{l|rrrr}
\toprule
   & $\Delta \text{NFR}_c$ & $\Delta_{\%}\text{NFR}_c\do$ &$\Delta \text{PFR}_c$ & $\Delta \text{acc}_c$ \\
\midrule
$\mathcal{L}_{comp}^{\mathcal{M}^{C}_{v2} \neq y}$ & \textbf{-3.97} & \textbf{-34.25} & \textbf{0.54} & \textbf{4.51} \\
$\mathcal{L}_{comp}^{\mathcal{M}_{v1} = y}$ & -2.29 & -19.76 & -0.60 & 1.68 \\
$a_{\mathcal{M}_{v1}}$ & -2.99 & -25.80 & -0.82 & 2.17 \\
CE+$\mathcal{L}_{comp}^{\mathcal{M}^{C}_{v2} \neq y}$ & -3.10 & -26.75 & 0.38 & 3.48 \\
CE+$\mathcal{L}_{comp}^{LL_L}$ & -2.12 & -18.29 & -0.49 & 1.63 \\
CE+$\mathcal{L}_{comp}^{LL_S}$ & -1.96 & -16.91 & 0.05 & 2.01 \\
\bottomrule
\end{tabular}
}
    \caption{A model update from Llama 1 $\rightarrow$ Llama 2 on PIQA. Different ablations and masking strategies and their impact on negative flips, positive flips, and accuracy improvement. }
    \label{tab:ablation}
\end{table}
Aligning to the old model only if it is already correct with a mask \revisionred{$\mathcal{L}_{comp}^{\mathcal{M}_{v1} = y}$ with $m_i = \mathbf{1}[\sigma(z_{\mathcal{M}_{v1},i})= y_i]$} or without masking with KL ($a_{\mathcal{M}_{v1}}$) leads to a small reduction in negative flips. 
Our approach that aligns to the old model if the current prediction is incorrect \revisionred{($\mathcal{L}_{comp}^{\mathcal{M}^{C}_{v2} \neq y}$)}, leads to the best reduction in negative flips while providing the biggest accuracy gain and increased positive flips (\cref{tab:ablation}). This strategy has the additional advantage that it takes into account extended inconsistent flips explained in \cref{fig:addedNFs}, as it aligns to $\mathcal{M}_{v1}$ whenever $\mathcal{M}_{v2}$ is incorrect. For this best-performing strategy, we see that including cross-entropy loss \revisionred{(CE+$\mathcal{L}_{comp}^{\mathcal{M}^{C}_{v2} \neq y}$)} does not lead to additional performance gains. 
\subsection{Behavior for Different Model Pairs}
\revisionred{A general point in the analysis of negative flips is its connection with accuracy: if accuracy is 100\%, the NFR must be 0\%. This means that compatible model updates become harder when updating $\mathcal{M}_{v1} \rightarrow \mathcal{M}_{v2}$ with lower $\mathcal{M}_{v2}$ accuracy.
We report observed compatibility when using vanilla adapter learning, and show a reduction in NFR when using the proposed method (MUSCLE). When $\text{acc}(\mathcal{M}_{v2}) > \text{acc}(\mathcal{M}_{v1})$ for a task, the observed NFR is small (\cref{motivation}). This has been the case irrespective of the model update scenario and task. We find that MUSCLE reduces the observed NFR the most when $\text{acc}(\mathcal{M}_{v2}) \approx \text{acc}(\mathcal{M}_{v1})$. In such cases, the incompatibility between vanilla model updates is not dominated by the fact that one model version is more accurate than the other, making it an easier use case to be improved by our proposed distillation loss. With multiple frequent incremental model updates in academia and industry, this is a very relevant use case to focus on.}

\section{Conclusion}
In this work, we study the task-specific compatibility problem when updating LLMs. We show that LLM updates with different scenarios, e.g., changes in model architecture, optimization, or training dataset, exhibit significant negative flips -- instances previously classified or generated correctly, and incorrectly after the model update.  We extend the negative flip metric for discriminative and, for the first time, generative tasks, and report results for various models and tasks.

We propose a novel method (MUSCLE) to train task-specific compatibility adapters when updating an old LLM to a new LLM to reduce negative flips while maintaining performance gain. Our proposed method does not require a modification to the base model training and is only based on adapter training. Further, as opposed to previous works, the proposed solution does not require both versions of the model in memory to enable compatibility, which is often infeasible due to the large size of LLMs.

We observe a mitigation of negative flips of 40\% for multiple-choice type evaluations, and 27\% for continuous summarization evaluation. We also show insights into model properties that facilitate transfer, finding that our alignment masking strategy provides best results with the additional benefit of mitigating inconsistent update behavior.

\section{Limitations, Risks and Future Work}
We do not consider a model update that includes changes in tokenization and/or vocabulary size (e.g.Llama 2 \cite{touvron2023llama2} to LLama 3 \cite{llama3modelcard}). Future work can explore compatible vocabulary mapping strategies before learning from prior model versions.

\paragraph{Tackling Large Performance Gaps}\revisionred{When there is a large performance gap between $M_{v1}$ and $M_{v2}$, loss hyperparameter weighting could be an interesting avenue to explore. We unsuccessfully experimented with simple weighting based on average accuracy. We hypothesize that instance-based loss weighting could be a more promising approach to tackle this model update case. As this would require intensive experimentation, we leave this as an avenue for future work.} In general, the utility of aligning to prior model versions is limited by the performance of the prior model version. For example, see the update from Phi 1 to Phi 1.5 in \cref{gsm8k}, where Phi 1 only has an accuracy of 3\%. In this case, it is arguable if an alignment to $\mathcal{M}_{v1}$ is desired and if the strive for compatibility outweighs a possible performance drop.

\paragraph{MUSCLE Performance Improvements}\revisionred{In our work, we observe an interesting overall performance improvement following knowledge transfer, which is an intriguing finding regarding the possibilities of this line of work. Previous works on model compatibility using distillation-like losses have also observed such a phenomenon (e.g., \cite{jaeckle2023fastfill}). Given that we initialize the compatibility adapter with $M_{v2}$ (an independently trained task adapter) and continue fine-tuning it with knowledge transfer loss from $M_{v1}$, one can argue that the observed performance improvement demonstrates an ensemble knowledge effect (i.e., knowledge from both $M_{v2}$ and $M_{v1}$ is aggregated into our compatibility adapter).}

\paragraph{Ethical Considerations and Risks}
One potential risk of the proposed approach for compatible task-specific LLM updates is the transfer of potential biases from the old model, $\mathcal{M}_{v1}$, to the new model trained through knowledge transfer. We did not explore this aspect in the current study.
\section{Acknowledgements}
We would like to thank Rick Chang, Cheng-Yu Hsieh and Yen-Ju Lu for their help with the paper.
\clearpage
\bibliography{emnlp2023}

\appendix

\section{Appendix}
\subsection{Training Hyperparameters and Evaluation}
We show an overview of the design choices of our LoRA adapter for task and compatibility training in \cref{tab:hyperparams}. We learn about the entire context and answer tokens during fine-tuning. We find that successfully training the compatibility adapter generally requires a larger LoRA rank. Task adapters can be trained with lower rank, but also experience performance boost with higher rank. We use the training splits of the respective datasets where we use 80\% of the training set as training data, and 20\% as validation data for selecting the best model based on validation loss. We evaluate for selection after each epoch. All compatibility adapter models are initialized with the previously trained task adapter models for version 2 model. KL divergence temperature is set to 2. We use deepspeed for optimization \cite{deepspeed}.
\begin{table}[h]
    \centering
    \small
    \begin{tabular}{l|l}
         Hyperparameter & Value\\
         \midrule
         Epochs & 10 \\
         Learning Rate & 1e-4 \\
         Grad. acc. steps & 8\\
         LoRA $\alpha$ & 256 \\
         LoRA rank & 128 \\
         Dropout & 0.0 \\
         Adapter Layers & All linear\\
         Warmup Steps & 500 \\
         Weight decay & 0.0 \\
    \end{tabular}
    \caption{Hyperparameters for the training setup of the task and compatibility adapters.}
    \label{tab:hyperparams}
\end{table}
For benchmark evaluation, we use LM evaluation harness \cite{eval-harness}, where most of the benchmarks (GSM8k, HellaSwag, PIQA) are already defined, and we use them as-is. We add an evaluation for SAMsum summarization evaluating ROUGE-1 score with a no-repeat N-gram size of 2, and generation stop words of ``Dialogue:'', ``Summary:'' (which are the keywords for the context and answer behavior), ``</s>'' and double new lines. All of our tasks are based on the English language. 

\subsection{Training Cost}
\revisionred{All Experiments were run on NVIDIA A100 and H100 GPUs with an overall compute budget of 720x8 GPUh. Assuming hourly rates of 2.5\$, this would amount to 14,400\$. Compared to regular parameter-efficient fine-tuning, our compatibility method requires both model versions in GPU memory during training, hence fewer data batches can be processed per time step. However, as the compatibility adapters are initialized with the task adapters, we can view it as a continued training. There are no differences for inference time and costs compared to regular parameter-efficiently fine-tuned LoRa adapters.}
\subsection{Model Update Evaluation}
In the evaluation of \revisionred{model update} regression for different model updates (\cref{motivation}), we consider the model pairs shown in \cref{tab:mp}.
\begin{table}[]
    \centering
    \small
    \begin{tabular}{l|l}
    $\mathcal{M}^{\text{base}}_{v1}$ &  $\mathcal{M}^{\text{base}}_{v2}$ \\
    \midrule
         Phi 1 & Phi 1.5 \\
Phi 1 & Phi 2 \\
Phi 1.5 & Phi 2 \\
Vicuna 1.3 & Vicuna 1.5 \\
Llama 1 & Llama 2 \\
Llama 1& Vicuna 1.5 \\
Llama 1 & Vicuna 1.3 \\
Vicuna 1.3 & Llama 2 \\
Llama 2 & Vicuna 1.5 \\
Llama 2 & Llama 3 \\
Llama 1 & Llama 3 \\
Vicuna 1.3 & Llama 3 \\
Vicuna 1.5 & Llama 3 \\
\bottomrule
\end{tabular}
    \caption{Model pairs used to analyse \revisionred{model update} regression.}
    \label{tab:mp}
\end{table}

\subsection{Auxiliary Cross-Entropy Loss}\label{celoss}
\revisionred{
For model updates with large performance gaps between $\mathcal{M}_{v1}$ and $\mathcal{M}_{v2}$, we find that an auxiliary cross-entropy loss enhances the stability of training. When the performance of $\mathcal{M}_{v1}$ is greatly inferior to $\mathcal{M}_{v2}$, we assume that the inferior $\mathcal{M}_{v1}$ introduces errors that steer the model away too much from the ground truth. We can account for this by adding the cross-entropy loss to align with the ground truth. Using \cref{eqn:loss}, we add a $\mathcal{L}_{CE}$ for training (scaled with hyperparameter $\lambda$) . }
\begin{equation}
\mathcal{L}_{CE} = -\frac{1}{N} \sum_{i=1}^N \sum_{k=1}^K y_{i,k} \log(\sigma(z_{\mathcal{M}^C_{v2},i,k}))
\end{equation}
\begin{equation}
\mathcal{L} = \lambda \mathcal{L}_{comp}^m + (1-\lambda) \mathcal{L}_{CE}
\end{equation}
\revisionred{
We find that likelihood-based masking (\cref{eqn:LLbased}) also requires auxiliary cross-entropy loss for stability. Even though one model version might provide a higher likelihood than the other, this likelihood does not necessarily mean a suitable probability distribution over the vocabulary, hence adding alignment to the ground truth helps to provide better results. 
}
\subsection{Downstream Tasks}
\begin{table}
\scalebox{0.95}{
\small
\begin{tabularx}{0.5\textwidth}{l|ll}
     Dataset & Train & Test   \\
     \midrule
     HellaSwag~\citep{zellers2019HellaSwag} & 40k & 10k\\
     PIQA~\citep{DBLP:journals/corr/abs-1911-11641} & 16k & 3k\\ 
     GSM8k~\citep{DBLP:journals/corr/abs-2110-14168} & 7.5k & 1.3k\\
     SAMsum~\citep{gliwa2019samsum} &  14.7k & 819\\ 
\end{tabularx}
}\caption{Dataset size for our experiments. }\label{tab:taskssize}
\end{table}
We describe the datasets used in our experiments. Each dataset poses unique challenges and targets different aspects of language understanding and reasoning. Dataset sizes are shown in \cref{tab:taskssize}.

\paragraph{GSM8K}
The Grade School Math 8K (GSM8K) dataset is designed to evaluate mathematical reasoning capabilities. It consists of grade school-level math word problems. These problems require the application of mathematical concepts and the ability to reason about quantities and operations in a textual format. The dataset is structured to test the performance of models on arithmetic, algebra, geometry, and statistics problems, reflecting a wide range of mathematical knowledge in education. We use this dataset as a representative case to test problem-solving cases. We evaluate \revisionred{model update} regression in this task with exact match, which is a strict evaluation on a fraction of produced tokens, but does not account for the reasoning process. 

\paragraph{SAMSum}
The SAMSum dataset consists of dialogue summaries designed to facilitate the evaluation of automatic conversational summarization models. It contains dialogue instances, paired with human-written summaries. These conversations mimic real-life scenarios to enable learning to generate coherent and concise summaries. We use this dataset as a representative case to test language generation behavior. We evaluate \revisionred{model update} regression in this task with ROUGE-1.

\paragraph{HellaSwag}
HellaSWAG is a dataset for assessing common sense reasoning and predictive text modeling. It builds on the SWAG~\cite{zellers2018swag} dataset by providing more challenging distractors. HellaSWAG consists of multiple-choice scenarios where a model must predict the most plausible continuation among four options, focusing on everyday activities, scenarios, and interactions. We use this dataset as a representative case to test abilities that require not just linguistic understanding but also real-world knowledge and common sense reasoning. We evaluate  \revisionred{model update} regression in this task with log-likelihoods for the correct answer, as multiple choices are given.

\paragraph{PIQA}
The Physical Interaction Question Answering (PIQA) dataset tests the understanding of physical and causal interactions in the real world through textual descriptions. It contains scenarios that require reasoning about physical properties, actions, and outcomes. Each scenario is presented as a question with two possible solutions, where the model must choose the most physically plausible one. We use this dataset as a representative case for evaluating models on tasks that require an understanding of the physical world and its governing principles.  We evaluate \revisionred{model update} regression in this task with log-likelihoods for the correct answer, as different choices are given.

\end{document}